\documentclass[
  dvipsnames,
  format=sigconf,
  nonacm
]{acmart}

\usepackage{amsmath}
\usepackage{mathtools}
\usepackage{booktabs}
\usepackage{multirow}
\usepackage{tabularx}
\usepackage{graphicx}
\usepackage[ruled,vlined]{algorithm2e}
\DontPrintSemicolon
\usepackage{xspace}
\usepackage{microtype}
\usepackage{url}

\setcopyright{none}

\title{Online Task Adaptation via Self-Organisation}

\author{Krsto Proroković}
\affiliation{%
  \institution{Independent Researcher}
  \country{Montenegro}
}
\email{krsto@speedwagon.ai}

\keywords{
meta-learning,
self-organisation,
neural cellular automata,
online adaptation,
learned plasticity
}

\begin{document}


\begin{abstract}
Neural networks are typically adapted by computing gradients and updating model parameters. We investigate whether task-specific adaptation can instead emerge from a meta-learned self-organising process that requires no gradients at adaptation time. We instantiate this idea with a Neural Cellular Automaton in which locally interacting recurrent cells maintain both a recurrent state and a fast associative memory. During meta-training, backpropagation is used to learn the recurrent dynamics together with how the memory is read and written. Once training is complete, the slow model parameters remain fixed, and online adaptation occurs only through cellwise memory updates driven by local prediction errors and a delta rule.

We evaluate whether the learned mechanism can adapt to semantically distinct held-out classification tasks. A single pass over the support data produces substantial improvements in held-out performance without gradient computation or parameter updates during adaptation, and the mechanism remains effective across large changes in the number of examples processed jointly. These results show that task-specific adaptation can be achieved through explicit fast-memory updates while keeping the slow model parameters fixed.

\end{abstract}

\maketitle

\section{Introduction}
\label{sec:introduction}

Artificial neural networks are typically trained by backpropagation. After a
forward computation produces a prediction, error information is propagated
backwards through the network to determine how its parameters should change.
Although highly effective, this creates a sharp distinction between inference
and learning: adaptation requires a separate gradient-based backward
computation through the model. Biological neural systems are not known to
implement backpropagation in this standard form \citep{lillicrap2020backpropagation}. They do, however, contain
extensive recurrent and feedback connections \citep{felleman1991distributed, markov2014anatomy, lillicrap2020backpropagation}, so the relevant distinction is
not whether information can travel backwards, but whether adaptation requires
gradients to be propagated through the computation that produced the
prediction.

We ask whether a model can learn to adapt from prediction errors without
backpropagating gradients through its computation during adaptation. We
instantiate this idea using a Neural Cellular Automaton (NCA)
\citep{mordvintsev2020growing}, in which locally interacting recurrent cells maintain both a recurrent state and a fast associative memory. During inference, cells
repeatedly exchange information and read from the fast memory to form
predictions. Once a target is observed, each cell computes a local prediction
error and uses it to update the memory using the delta rule \citep{widrow1960adaptive}. No gradients are computed and no network parameters are modified during adaptation.
We view the resulting adaptation process as a form of self-organisation: recurrent interactions and cellwise memory updates collectively give rise to task-specific behaviour.

While we rely on backpropagation for meta-learning, the resulting adaptation algorithm requires no gradients. Meta-training learns the recurrent dynamics and how the fast memory is read and written; once training is complete, the slow model parameters remain fixed and online adaptation to a new task occurs only through changes to the fast memory.

We evaluate the resulting mechanism on five-class classification tasks derived
from the superclass structure of CIFAR-100, with training and held-out tasks
drawn from disjoint superclasses. Starting from an empty memory, a single pass
over the support set raises mean held-out accuracy from \(20.0\%\) to
\(48.2\%\), compared with \(54.4\%\) for the corresponding baseline trained
from scratch on each task using backpropagation. The learned adaptation
mechanism also remains effective when the number of support examples processed
jointly is changed substantially. These results show that substantial
task-specific adaptation can be achieved through an explicit fast memory whose
updates require no gradient computation at adaptation time.

\section{Method}
\label{sec:method}

We consider image classification with inputs \(x\in\mathbb{R}^{H\times W\times C_x}\) and labels \(k\in\{1,\ldots,K\}\). Although we present the method in this setting, it can be adapted to other input structures, such as vectors or voxels, and to other supervised adaptation tasks for which target outputs are available.

In this work, we implement the method with an NCA, but the same idea could be applied to other architectures that iteratively update a set of interacting units using shared parameters, including Universal Transformers \citep{dehghani2019universal}, looped Transformers \citep{giannou2023looped}, and recurrent graph neural networks \citep{li2016gated}.

\subsection{Inference}
\label{subsec:inference}

Each cell \((i,j)\) maintains a recurrent state
\(s_{ij}(t)\in\mathbb{R}^{C_s}\) and a fast associative memory
\(M_{ij}\in\mathbb{R}^{C_v\times C_k}\). We refer to \(s(t)\in\mathbb{R}^{H\times W\times C_s}\) as the state at recurrent step \(t\), and to \(M\in\mathbb{R}^{H\times W\times C_v\times C_k}\) as the memory. During inference, the memory remains fixed while the recurrent state evolves
over \(T\) steps.

The recurrent state is initialised as
\[
s(0)=0,
\]
and the memory is initially set to
\[
M=0.
\]
In principle, the initial memory could instead be learned directly or generated by a parameterised function from a compact set of parameters (e.g. using \citep{sitzmann2020implicit, mordvintsev2020growing, mordvintsev2022growing}). This would allow adaptation to begin from a learned prior rather than an empty memory, analogous to biological development being shaped by inherited regulatory structure rather than beginning from a blank state \citep{davidson2010emerging}. We leave such learned or generated memory initialisation to future work.

To read from memory, the cell projects its current state into a \(C_k\)-dimensional key,
\begin{equation*}
k^{\mathrm{read}}_{ij}(t)
=
\operatorname{norm}\!\left(W^{\mathrm{read}}s_{ij}(t)\right),
\end{equation*}
where \(W^{\mathrm{read}}\in\mathbb{R}^{C_k\times C_s}\) is learned and \(\operatorname{norm}(\cdot)\) denotes \(L_2\) normalisation. The memory readout is
\begin{equation*}
r_{ij}(t)
=
M_{ij}k^{\mathrm{read}}_{ij}(t)
\in\mathbb{R}^{C_v}.
\end{equation*}

Cells communicate by perceiving the recurrent states of neighbouring cells.
Let \(\mathcal{N}_{ijc}(t)\) denote the spatial neighbourhood of channel \(c\)
centred at position \((i,j)\) at step \(t\). Each channel has its own bank of \(F\) learnable filters \(\kappa_{cf}\), unlike the original NCA formulation in which filters are shared across channels~\citep{mordvintsev2020growing}. The
resulting perception vector is
\begin{equation*}
p_{ij}(t)
=
\left[
    \left[
        \left\langle
        \kappa_{cf},
        \mathcal{N}_{ijc}(t)
        \right\rangle
    \right]_{f=1}^{F}
\right]_{c=1}^{C_s}
\in\mathbb{R}^{F C_s},
\end{equation*}
where \(\langle\cdot,\cdot\rangle\) denotes the inner product and
\([\cdot]\) denotes concatenation.

The update module is a single-hidden-layer multilayer perceptron (MLP)
applied independently at each spatial position. For each cell \((i,j)\),
we compute
\begin{equation*}
\Delta s_{ij}(t) =
W_\Delta
\operatorname{ReLU}
\left(
W_h
\left[
s_{ij}(t),
p_{ij}(t),
r_{ij}(t),
x_{ij}
\right]
+
b_h
\right),
\end{equation*}
where \(x_{ij}\) is the input feature vector associated with cell \((i,j)\). The input \(x_{ij}\) is clamped throughout inference and is
provided to the update module at every recurrent step. Thus, while the
recurrent state evolves over time, each cell retains direct access to the
corresponding input features.

The state is then updated residually,
\begin{equation*}
s(t+1)=s(t)+\xi(t)\odot\Delta s(t),
\end{equation*}
where \(\xi(t)\) is a stochastic firing mask.
During meta-training, the update mask \(\xi(t)\) is constructed by sampling
one independent Bernoulli variable with firing rate \(\rho\) for each spatial
position and recurrent step, then broadcasting this scalar mask across all
\(C_s\) state channels. During meta-validation and meta-testing, including adaptation on the support set, we replace the stochastic mask by its expectation, setting the scalar mask value to \(\rho\) everywhere, analogous to replacing stochastic dropout by its expected activation at evaluation time \citep{srivastava2014dropout}.

Each cell produces a local output from its final recurrent state,
\begin{equation*}
\hat y_{ij}
=
W_{\mathrm y}s_{ij}(T)+b_{\mathrm y},
\end{equation*}
where \(\hat y_{ij}\in\mathbb{R}^{C_y}\). For classification, \(C_y = K\), and the final prediction is obtained by averaging the
cellwise class probabilities and selecting the most probable class,
\begin{equation*}
\hat k
=
\arg\max_c
\frac{1}{HW}
\sum_{i=1}^{H}
\sum_{j=1}^{W}
\operatorname{softmax}\!\left(\hat y_{ij}\right)_c.
\end{equation*}

\subsection{Adaptation}
\label{subsec:adaptation}

For each cell, we compute an error signal by comparing its output
\(\hat y_{ij}\) with the target \(y\). For classification, 
\(y\in\mathbb{R}^{K}\) is the one-hot encoding of the correct class \(k\),
optionally label-smoothed \citep{szegedy2016rethinking}.
We define the cellwise error as
\begin{equation*}
e_{ij}
=
\operatorname{softmax}\!\left(\hat y_{ij}\right)-y.
\end{equation*}

To update the memory, each cell computes a write key from its final recurrent
state and a write value from its final recurrent state and error signal.
The write key is obtained through a learned linear projection,
\begin{equation*}
k^{\mathrm{write}}_{ij}
=
\operatorname{norm}\!\left(
W^{\mathrm{write}}s_{ij}(T)
\right)
\in\mathbb{R}^{C_k}.
\end{equation*}
The write value is produced by a single-hidden-layer MLP,
\begin{equation*}
v_{ij}
=
W_v^{(2)}
\operatorname{ReLU}\!\left(
W_v^{(1)}
\left[
s_{ij}(T),e_{ij}
\right]
+
b_v^{(1)}
\right)
\in\mathbb{R}^{C_v}.
\end{equation*}

Let \(\gamma\) denote the maximum possible \(L_2\)-norm of the cellwise
error. We define the write strength as
\begin{equation*}
\eta_{ij}
=
\frac{\lVert e_{ij}\rVert_2}{\gamma},
\end{equation*}
which ensures \(\eta_{ij}\in[0,1]\) and \(\eta_{ij}=0\) whenever
\(e_{ij}=0\). For classification, \(\gamma\) can be computed analytically
because both the softmax output and the (possibly label-smoothed) one-hot
target lie in bounded subsets of \(\mathbb{R}^{K}\).

The memory is updated using the delta rule \citep{widrow1960adaptive, schlag2021linear}. For a batch of
\(B'\) examples, the individual updates are summed and scaled by \(1/B\),
where \(B\) is the batch size used during meta-training:
\begin{equation*}
M_{ij}
\leftarrow
M_{ij}
+
\frac{1}{B}
\sum_{b=1}^{B'}
\eta_{bij}
\left(
v_{bij}
-
M_{ij}k_{bij}^\mathrm{write}
\right)
\left(k_{bij}^\mathrm{write}\right)^\top.
\end{equation*}
Each term moves the value retrieved at the corresponding write key toward
\(v_{bij}\), with strength determined by \(\eta_{bij}\).
Equivalently, the
update is the mean per-example update over the current batch, multiplied by
\(B'/B\). Thus, each example contributes with the same \(1/B\) scaling used
during meta-training.
We restrict \(B' \leq B\); larger batches are split into batches of at most
\(B\) examples and processed sequentially.

\subsection{Meta-training}
\label{subsec:meta-training}

Backpropagation is used during meta-training to optimise the slow parameters; the adaptation rule itself consists solely of the memory updates described above.
During meta-training, the model is trained on a meta-dataset consisting of
multiple tasks. Each task is associated with a dataset of labelled examples,
which is presented to the model as a sequence of mini-batches of size \(B\).
Unlike conventional episodic meta-learning \citep{finn2017model, snell2017prototypical}, this dataset is not split into separate support and query sets. 
The same sequence of labelled examples is used both to provide the meta-training objective and to update the memory.
For each batch, the recurrent state is reset, the model performs inference
using its current memory, and only then is the memory updated using the
adaptation rule described above. The updated memory is carried forward and
used when processing the subsequent batch.

The slow model parameters are optimised by backpropagation through windows of
\(L\geq2\) consecutive batches. The first batch of each task is used only for
adaptation and does not contribute to the meta-training loss; all subsequently processed batches contribute to the loss. For each contributing batch, the meta-training
loss is the cross-entropy between the target and the output of each cell,
averaged over spatial positions and examples in the batch. Applying the loss
after every batch encourages the adaptation mechanism to improve performance
online throughout the task, rather than only after the full sequence of
updates has been observed. After each window, the slow parameters are updated
using gradient-based optimisation. At window boundaries, the current memory
is carried forward, but gradients are not propagated through preceding
windows.

The model processes each task dataset in this manner. At the beginning of a new task, the memory is reset. We refer to one
complete pass through the meta-training task set as a meta-epoch. The complete
meta-training procedure is summarised in Algorithm~\ref{alg:meta-training}.

\begin{algorithm}[t]
\caption{Meta-training}
\label{alg:meta-training}

\Repeat{stopping criterion is met}{
    \tcp{one meta-epoch}
    \For{each task}{
        reset memory \(M\)\;
        \tcp{adaptation-only batch}
        take the first mini-batch \((x,y)\)\;
        reset recurrent state\;
        \(\hat y \gets \mathrm{Inference}(x,M)\)\;
        \(M \gets \mathrm{Adapt}(M,\hat y,y)\)\;

        \For{each window of \(L\) mini-batches}{
            \(\mathcal{L} \gets 0\)\;

            \For{each mini-batch \((x,y)\) in the window}{
                reset recurrent state\;
                \(\hat y \gets \mathrm{Inference}(x,M)\)\;
                \(\mathcal{L} \gets
                \mathcal{L} + \mathrm{Loss}(\hat y,y)\)\;
                \(M \gets \mathrm{Adapt}(M,\hat y,y)\)\;
            }

            update slow parameters using \(\nabla \mathcal{L}\)\;
            \tcp{truncate backpropagation}
            \(M \gets \operatorname{stopgrad}(M)\)\;
        }
    }
}
\end{algorithm}

\section{Related work}
\label{subsec:related-work}

A large class of meta-learning methods performs adaptation by modifying the parameters of a base learner. Methods such as Model-Agnostic Meta-Learning \citep{finn2017model} and its variants \citep{li2017meta, zintgraf2019fast, rajeswaran2019meta, raghu2020rapid} meta-learn parameters that can be rapidly adapted to a new task using gradient-based optimisation. Learned optimisers such as the meta-learner LSTM \citep{ravi2017optimization} additionally meta-learn the parameter update rule, while TURTLE \citep{huisman2022stateless} jointly learns the initial parameters and a stateless neural update rule. Related approaches based on in-context learning adapt to new tasks without modifying model parameters by conditioning predictions on demonstrations from the current task \citep{brown2020language}. In contrast, our method accumulates task-specific information in an explicit fast associative memory, which acts as a task context that is updated
incrementally across observations.

A related line of work meta-learns mechanisms for rapid adaptation through learned plasticity. Differentiable Plasticity \citep{miconi2018differentiable} jointly learns conventional connection weights and the plasticity of selected connections, allowing recurrent networks to adapt through Hebbian updates after training. Backpropamine \citep{miconi2019backpropamine} extends this approach by learning neuromodulatory signals that control when and how plasticity occurs. Other work has meta-learned local plasticity rules together with feedback pathways for online learning \citep{shervani2023meta}. More generally, Variable Shared Meta Learning \citep{kirsch2021meta} meta-learns learning algorithms within shared recurrent dynamics and can perform adaptation without explicit gradient computation. Our method shares the goal of learning an adaptation mechanism that does not
require gradients at meta-test time, while imposing an explicit separation
between slow model parameters and fast task-specific memory. Adaptation
modifies only the cellwise memory \(M\) according to a prescribed local update
rule, while the slow network parameters remain fixed.

Our method is most closely related to fast-weight and self-\linebreak[4]modifying neural networks. Early fast-weight programmers and subsequent fast-weight memory models separated slowly learned parameters from rapidly changing weights used as temporary memory \citep{schmidhuber1992learning, ba2016using}. Memory-Augmented Neural Networks \citep{santoro2016meta} similarly use a separate rapidly updated memory to support few-shot learning, while Meta Networks \citep{munkhdalai2017meta} combine slow and fast parameters for rapid task adaptation by generating task-dependent fast weights from gradient-based meta-information extracted from labelled support examples. More recently, the Self-Referential Weight Matrix (SRWM) \citep{irie2022modern} introduced a scalable mechanism in which a weight matrix modifies itself at runtime using outer-product and delta-rule updates, and this framework was subsequently extended to meta-learn continual learning algorithms \citep{irie2025metalearning}. Like these approaches, our method separates slow meta-learned parameters from fast task-dependent adaptation. However, rather than allowing a single general weight matrix to modify itself, we restrict adaptation to a spatially distributed associative memory, with a separate memory matrix maintained by each locally interacting cell. The memory is updated by a prescribed delta rule, while the recurrent dynamics meta-learn the read and write representations. The write strength is explicitly determined by the magnitude of the prediction
error, such that zero error necessarily results in no memory modification,
whereas SRWM learns its write strength as part of the self-modification
mechanism. SRWM updates its fast weights sequentially as observations are processed,
whereas our method can process all examples in a batch in parallel, aggregate
their memory updates, and apply the resulting update jointly. This batch-level formulation naturally supports changes in the number of examples processed jointly during adaptation. SRWM and its continual-learning extension also use separate examples for fast adaptation and for defining the meta-training objective, whereas our meta-training procedure uses the same online stream both to update the fast memory and to provide the meta-training loss.

\section{Experiments}
\label{sec:experiments}

\subsection{Experimental setup}

\paragraph{Dataset and tasks.}
We evaluate on CIFAR-100, whose 100 classes are grouped into 20 superclasses
of five classes each. A task corresponds to one superclass and therefore
defines a five-class classification problem among related classes, for example
the five kinds of fish. Labels are remapped to \(\{1,\ldots,5\}\) within
each task, so the label space is shared across tasks and a class index does
not identify which task is being solved.

We adopt the superclass partition of FC100 \citep{oreshkin2018}, which divides the
20 superclasses into 12 meta-training, 4 meta-validation, and 4 meta-test
superclasses. Splitting by superclass rather than by class prevents
semantically related classes from straddling the meta-learning boundary.
For example, because \texttt{motorcycle} and \texttt{bicycle} belong to the
same superclass, a meta-test task cannot contain one while a meta-training
task contains the other. The resulting partition also induces a substantial
semantic shift between meta-training and evaluation tasks: meta-training
contains one reptile and one fish superclass and is otherwise dominated by
objects, plants, vehicles, and scenes, whereas the meta-validation and
meta-test tasks are predominantly mammals and invertebrates. The complete
partition is listed in Table~\ref{tab:fc100-partition}.

\begin{table}[t]
\centering
\caption{FC100 superclass partition. We reproduce it here because the partition is reported only in the supplementary material of the original paper.}
\label{tab:fc100-partition}
\begin{tabularx}{\linewidth}{lX}
\toprule
\textbf{Split} & \textbf{Superclasses} \\
\midrule
Meta-train (12) &
fish, flowers, food containers, fruit and vegetables,
household electrical devices, household furniture,
large man-made outdoor things, large natural outdoor scenes,
reptiles, trees, vehicles 1, vehicles 2 \\
\addlinespace
Meta-val (4) &
large carnivores, large omnivores and herbivores,
non-insect invertebrates, small mammals \\
\addlinespace
Meta-test (4) &
aquatic mammals, insects, medium-sized mammals, people \\
\bottomrule
\end{tabularx}
\end{table}

We use the FC100 class partition but not its task-sampling protocol. In FC100, tasks are formed by sampling classes from the corresponding meta-training, meta-validation, or meta-test class pool, so an individual task will typically contain classes drawn from several superclasses. In contrast, our tasks are the superclasses themselves, making each task a fine-grained discrimination problem within a single semantic group---for example, distinguishing five kinds of fish rather than distinguishing a fish from a chair. Our results are therefore not directly comparable to published FC100 results.

\paragraph{Protocol.}
Each superclass contains 3000 images: 2500 from the CIFAR-100 training split
and 500 from its test split. For meta-training tasks, all 3000 images are included in the task dataset.
Because the meta-training and meta-test tasks contain disjoint superclasses,
using both CIFAR-100 splits for meta-training tasks does not expose the model
to any meta-test classes. During meta-training, the model predicts on each
batch before adapting on it, so every prediction is made on examples that have
not previously contributed to the memory. The same stream can therefore
provide both the meta-training objective and the memory updates.

At meta-validation and meta-test time, the model adapts over the 2500
training-split images and is then evaluated on the 500 test-split images with
the memory frozen. Using the original CIFAR-100 split rather than a random
split makes the evaluation set reproducible without reference to a seed.
Support data is presented in a fixed order during evaluation so that metrics are comparable across checkpoints and configurations.

No data augmentation is used. Because augmentation would alter the examples
written to memory and thereby change the adaptation trajectory, we evaluate on
the original images and use the same input distribution during meta-training.

\paragraph{Model architecture.}
Rather than operating directly on the \(32\times32\) input image, the NCA
receives features produced by a small convolutional backbone. The backbone
consists of two \(3\times3\) convolutions with stride \(2\), mapping \(3\)
input channels to \(32\) and then \(64\) feature channels, with a ReLU
activation after each convolution. This reduces the spatial resolution from
\(32\times32\) to \(8\times8\), substantially reducing the computational cost
of the recurrent dynamics and shortening the spatial distances over which
information must propagate.

We use \(C_s=C_k=C_v=32\) and \(F=3\) learnable \(3\times3\) perception filters
per state channel. The input to the NCA update module therefore has dimension
\begin{equation*}
C_s + FC_s + C_v + 64
=
32 + 3\cdot32 + 32 + 64
=
224.
\end{equation*}

The hidden layer of the update MLP has width \(448\), twice its input
dimension. For the value module, the input dimension is
\(C_s+C_y=32+5=37\), and the hidden dimension is \(74\). The complete model
contains \(141{,}793\) trainable parameters. The NCA is unrolled for \(T=16\)
recurrent steps with firing rate \(\rho=0.5\).

\paragraph{Baseline.}
As a baseline, we use the same convolutional backbone, perception filters,
update module, and output projection, but remove the read and write modules
and the fast associative memory. The update module retains the same hidden
width as in the proposed model, and the baseline uses the same recurrent
horizon \(T=16\) and firing rate \(\rho=0.5\). The baseline contains
\(121{,}221\) trainable parameters. For each task in the meta-test partition, it is
trained on the corresponding support set using gradient-based optimisation
and then evaluated on the task's query set.

\paragraph{Initialisation.}
The proposed model and the baseline use the same initialisation for all shared components. The backbone weights, perception filters \(\kappa_{cf}\), and output projection \(W_{\mathrm y}\) are initialised using Kaiming uniform initialisation \citep{he2015delving}. The read and write projections \(W^{\mathrm{read}}\) and \(W^{\mathrm{write}}\) are initialised the same way. The update-module matrix \(W_h\) is initialised using Kaiming uniform initialisation with the ReLU gain, while \(W_\Delta\) is initialised to zero. Similarly, \(W_v^{(1)}\) is initialised using Kaiming uniform initialisation with the ReLU gain, while \(W_v^{(2)}\) is initialised to zero. The biases \(b_h\), \(b_{\mathrm y}\), and \(b_v^{(1)}\) are initialised to zero.

\paragraph{Training.}
We use cross-entropy loss with label smoothing \(\alpha=0.1\). For \(K=5\),
this gives a maximum cellwise error norm of \(\gamma\approx1.345\), which
is used to normalise the write strength. Both the proposed model and the
baseline use a batch size of \(128\) and are optimised with AdamW \citep{loshchilov2018decoupled} using a
learning rate of \(10^{-3}\) and weight decay \(0.1\), applied uniformly to
all trainable parameters. Gradients are clipped to a maximum global
\(L_2\)-norm of \(1\).

At the beginning of each meta-epoch, the order of the meta-training tasks is
shuffled, and the examples within each task are shuffled before batching.
During meta-training, gradients are propagated through windows of \(L=8\)
consecutive batches. After the initial adaptation-only batch, only complete
windows of \(L\) loss-contributing batches are processed; any remaining
incomplete window at the end of the task is discarded. Meta-validation is
performed by adapting on the support set of each meta-validation task and
measuring the loss on its query set.

For the baseline, \(10\%\) of each support set is selected at random and
reserved for validation. The remaining examples are shuffled during training
and processed in complete batches of \(128\), with any final incomplete batch
discarded.

The learning rate is halved when the corresponding validation loss has not
improved for \(20\) epochs, or meta-epochs in the meta-learning setting.
Training is stopped after \(40\) epochs without improvement, using the same
convention for meta-training.

\subsection{Results}

We refer to each complete presentation of the support set during adaptation
(during which only the fast memory is modified) as one pass. After a single
pass, mean accuracy on held-out images rises from \(20.0\%\) to \(48.2\%\)
(Table~\ref{tab:results}). The lower figure is obtained by evaluating the same
model with its memory left empty, in which case performance is approximately
at chance on every task, ranging from \(17.9\%\) to \(21.9\%\) against a chance
level of \(20\%\). Because the slow parameters are unchanged between the empty-memory and adapted evaluations, the improvement arises entirely from the fast memory.

\begin{table}[t]
\centering
\caption{Accuracy on the 500 held-out images of each meta-test superclass.
Chance accuracy is \(20\%\). \emph{Empty} evaluates the model with the fast memory left at its zero initialisation; \emph{Adapted} evaluates the same model after a single pass over the 2500 support images, with the memory subsequently frozen. \emph{Scratch} uses the corresponding baseline architecture without fast memory and is trained on the support images using gradient-based optimisation.
Results report the mean and standard error over five seeds.}
\label{tab:results}
\begin{tabular}{lccc}
\toprule
Superclass & Empty & Adapted & Scratch \\
\midrule
Aquatic mammals      & 17.9 & 45.2 {\scriptsize$\pm$0.5} & 52.8 {\scriptsize$\pm$0.7} \\
Insects              & 21.9 & 59.8 {\scriptsize$\pm$0.8} & 63.9 {\scriptsize$\pm$0.5} \\
Medium-sized mammals & 19.1 & 54.6 {\scriptsize$\pm$0.8} & 64.8 {\scriptsize$\pm$0.7} \\
People               & 21.1 & 33.1 {\scriptsize$\pm$1.1} & 36.2 {\scriptsize$\pm$0.6} \\
\midrule
Mean                 & 20.0 & 48.2 {\scriptsize$\pm$0.3} & 54.4 {\scriptsize$\pm$0.2} \\
\bottomrule
\end{tabular}
\end{table}

Training the corresponding baseline architecture from scratch on each
superclass, with the fast memory and its read/write modules removed, reaches a
mean accuracy of \(54.4\%\). Fast-memory adaptation therefore recovers
approximately \(82\%\) of the gap between chance and the from-scratch
baseline, despite a substantial asymmetry in the adaptation available to the
two methods: the baseline makes tens of passes over the
support set with backpropagation, reserves \(10\%\) of it for early stopping,
and adapts its convolutional layers along with the rest of the network,
whereas the proposed model sees each support image once, computes no gradients
during adaptation, and modifies only its fast memory.

The meta-training set contains only twelve tasks, and the held-out superclasses are semantically different from the meta-training set. That the learned adaptation mechanism transfers under these conditions suggests that it generalises beyond the specific structure of the meta-training tasks.

The adaptation mechanism is also insensitive to the batch size used on the
support set. Because each update sums the per-example contributions with a
fixed scaling determined by the meta-training batch size, changing the number
of examples processed jointly does not directly change the contribution of
each example. Grouping does affect how often the correction term is recomputed,
since each write is made against the memory left by the previous one, but the
accuracies show that this has little effect in practice. Presenting the same
\(2500\) support images in batches of \(1\), \(16\), \(32\), \(64\), and
\(128\)---changing the number of memory writes from \(2500\) to \(20\)---yields
mean accuracies of \(48.3\%\), \(48.3\%\), \(48.4\%\), \(48.1\%\), and
\(48.2\%\), respectively.

Both methods perform worst on \emph{people}, reaching \(33.1\%\) and \(36.2\%\) accuracy for fast-memory adaptation and training from scratch, respectively. Its five classes are baby, boy, girl, man, and woman, so the task requires distinguishing both age and sex from \(32\times32\) images. That the from-scratch baseline reaches only \(36.2\%\) despite training on \(2250\) labelled examples suggests that much of the difficulty reflects the task itself under the chosen input resolution and architecture, rather than being specific to the proposed adaptation rule.

\subsection{Effect of truncation window across adaptation passes}

\begin{figure}[t]
    \centering
    \includegraphics[width=\columnwidth]{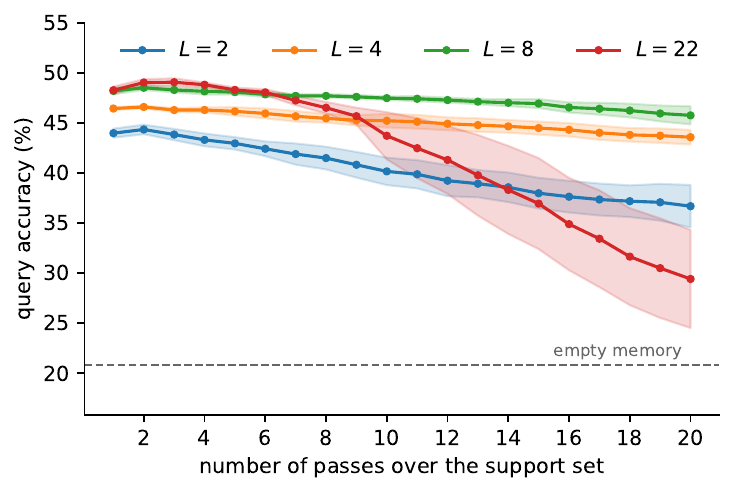}
    \caption{Query accuracy as a function of the number of passes over the support set for different truncation windows \(L\). The dashed line shows performance with the fast memory left at its zero initialisation. Shaded regions indicate the standard error over five seeds.}
    \label{fig:passes}
\end{figure}

Adaptation is not restricted to a single pass: because the memory persists, the
support set can be presented repeatedly. Figure~\ref{fig:passes} shows query
accuracy as a function of the number of passes for different truncation windows
\(L\), with the support set presented in batches of \(128\) throughout.

For \(L=4\) and \(L=8\), accuracy peaks on the second pass and then declines
slowly, losing fewer than three percentage points over the following eighteen
passes. A single pass therefore captures most of the available improvement, in
contrast to the from-scratch baseline, which requires tens of passes over the
same data. \(L=2\) performs worse throughout and declines more strongly, losing
\(7.4\) percentage points over repeated adaptation. With \(L=22\), which spans
all loss-contributing batches in a meta-training pass and therefore removes
truncation within the pass, accuracy is marginally higher over the first few
passes but does not remain so: it stays approximately stable until the ninth
pass and then falls sharply, reaching \(29.4\%\) by the twentieth. Full 
backpropagation through the pass therefore does not reduce initial adaptation 
performance in this setting, but produces the least stable behaviour under repeated memory updates.


\bibliographystyle{ACM-Reference-Format}
\bibliography{references}

@article{ba2016using,
  title={Using fast weights to attend to the recent past},
  author={Ba, Jimmy and Hinton, Geoffrey E and Mnih, Volodymyr and Leibo, Joel Z and Ionescu, Catalin},
  journal={Advances in neural information processing systems},
  volume={29},
  year={2016}
}

@article{brown2020language,
  title={Language models are few-shot learners},
  author={Brown, Tom and Mann, Benjamin and Ryder, Nick and Subbiah, Melanie and Kaplan, Jared D and Dhariwal, Prafulla and Neelakantan, Arvind and Shyam, Pranav and Sastry, Girish and Askell, Amanda and others},
  journal={Advances in neural information processing systems},
  volume={33},
  pages={1877--1901},
  year={2020}
}

@article{davidson2010emerging,
  title={Emerging properties of animal gene regulatory networks},
  author={Davidson, Eric H},
  journal={Nature},
  volume={468},
  number={7326},
  pages={911--920},
  year={2010},
  publisher={Nature Publishing Group UK London}
}

@inproceedings{dehghani2019universal,
  title={Universal Transformers},
  author={Mostafa Dehghani and Stephan Gouws and Oriol Vinyals and Jakob Uszkoreit and Lukasz Kaiser},
  booktitle={International Conference on Learning Representations},
  year={2019},
  url={https://openreview.net/forum?id=HyzdRiR9Y7},
}

@article{felleman1991distributed,
  title={Distributed hierarchical processing in the primate cerebral cortex.},
  author={Felleman, Daniel J and Van Essen, David C},
  journal={Cerebral cortex (New York, NY: 1991)},
  volume={1},
  number={1},
  pages={1--47},
  year={1991}
}

@inproceedings{finn2017model,
  title={Model-agnostic meta-learning for fast adaptation of deep networks},
  author={Finn, Chelsea and Abbeel, Pieter and Levine, Sergey},
  booktitle={International conference on machine learning},
  pages={1126--1135},
  year={2017},
  organization={PMLR}
}

@inproceedings{giannou2023looped,
  title={Looped transformers as programmable computers},
  author={Giannou, Angeliki and Rajput, Shashank and Sohn, Jy-yong and Lee, Kangwook and Lee, Jason D and Papailiopoulos, Dimitris},
  booktitle={International Conference on Machine Learning},
  pages={11398--11442},
  year={2023},
  organization={PMLR}
}

@inproceedings{he2015delving,
  title={Delving deep into rectifiers: Surpassing human-level performance on imagenet classification},
  author={He, Kaiming and Zhang, Xiangyu and Ren, Shaoqing and Sun, Jian},
  booktitle={Proceedings of the IEEE international conference on computer vision},
  pages={1026--1034},
  year={2015}
}

@article{huisman2022stateless,
  title={Stateless neural meta-learning using second-order gradients},
  author={Huisman, Mike and Plaat, Aske and van Rijn, Jan N},
  journal={Machine Learning},
  volume={111},
  number={9},
  pages={3227--3244},
  year={2022},
  publisher={Springer}
}

@inproceedings{irie2022modern,
  title={A modern self-referential weight matrix that learns to modify itself},
  author={Irie, Kazuki and Schlag, Imanol and Csord{\'a}s, R{\'o}bert and Schmidhuber, J{\"u}rgen},
  booktitle={International conference on machine learning},
  pages={9660--9677},
  year={2022},
  organization={PMLR}
}

@article{irie2025metalearning,
  title={Metalearning Continual Learning Algorithms},
  author={Kazuki Irie and R{\'o}bert Csord{\'a}s and J{\"u}rgen Schmidhuber},
  journal={Transactions on Machine Learning Research},
  issn={2835-8856},
  year={2025},
  url={https://openreview.net/forum?id=IaUh7CSD3k},
  note={}
}

@article{kirsch2021meta,
  title={Meta learning backpropagation and improving it},
  author={Kirsch, Louis and Schmidhuber, J{\"u}rgen},
  journal={Advances in Neural Information Processing Systems},
  volume={34},
  pages={14122--14134},
  year={2021}
}

@inproceedings{li2016gated,
  title={Gated graph sequence neural networks},
  author={Li, Yujia and Tarlow, Daniel and Brockschmidt, Marc and Zemel, Richard},
  booktitle={International Conference on Learning Representations},
  year={2016},
}

@article{li2017meta,
  title={Meta-sgd: Learning to learn quickly for few-shot learning},
  author={Li, Zhenguo and Zhou, Fengwei and Chen, Fei and Li, Hang},
  journal={arXiv preprint arXiv:1707.09835},
  year={2017}
}

@article{lillicrap2020backpropagation,
  title={Backpropagation and the brain},
  author={Lillicrap, Timothy P and Santoro, Adam and Marris, Luke and Akerman, Colin J and Hinton, Geoffrey},
  journal={Nature Reviews Neuroscience},
  volume={21},
  number={6},
  pages={335--346},
  year={2020},
  publisher={Nature Publishing Group UK London}
}

@inproceedings{loshchilov2018decoupled,
  title={Decoupled Weight Decay Regularization},
  author={Ilya Loshchilov and Frank Hutter},
  booktitle={International Conference on Learning Representations},
  year={2019},
  url={https://openreview.net/forum?id=Bkg6RiCqY7},
}

@article{markov2014anatomy,
  title={Anatomy of hierarchy: feedforward and feedback pathways in macaque visual cortex},
  author={Markov, Nikola T and Vezoli, Julien and Chameau, Pascal and Falchier, Arnaud and Quilodran, Ren{\'e} and Huissoud, Cyril and Lamy, Camille and Misery, Pierre and Giroud, Pascale and Ullman, Shimon and others},
  journal={Journal of comparative neurology},
  volume={522},
  number={1},
  pages={225--259},
  year={2014},
  publisher={Wiley Online Library}
}

@inproceedings{miconi2018differentiable,
  title={Differentiable plasticity: training plastic neural networks with backpropagation},
  author={Miconi, Thomas and Stanley, Kenneth and Clune, Jeff},
  booktitle={International conference on machine learning},
  pages={3559--3568},
  year={2018},
  organization={PMLR}
}

@inproceedings{miconi2019backpropamine,
  title={Backpropamine: training self-modifying neural networks with differentiable neuromodulated plasticity},
  author={Thomas Miconi and Aditya Rawal and Jeff Clune and Kenneth O. Stanley},
  booktitle={International Conference on Learning Representations},
  year={2019},
  url={https://openreview.net/forum?id=r1lrAiA5Ym},
}

@article{mordvintsev2020growing,
  title={Growing neural cellular automata},
  author={Mordvintsev, Alexander and Randazzo, Ettore and Niklasson, Eyvind and Levin, Michael},
  journal={Distill},
  volume={5},
  number={2},
  pages={e23},
  year={2020}
}

@inproceedings{mordvintsev2022growing,
  title={Growing isotropic neural cellular automata},
  author={Mordvintsev, Alexander and Randazzo, Ettore and Fouts, Craig},
  booktitle={Artificial Life Conference Proceedings 34},
  volume={2022},
  number={1},
  pages={65},
  year={2022},
  organization={MIT Press One Rogers Street, Cambridge, MA 02142-1209, USA journals-info~…}
}

@inproceedings{munkhdalai2017meta,
  title={Meta networks},
  author={Munkhdalai, Tsendsuren and Yu, Hong},
  booktitle={International conference on machine learning},
  pages={2554--2563},
  year={2017},
  organization={PMLR}
}

@inproceedings{oreshkin2018,
  title={TADAM: Task dependent adaptive metric for improved few-shot learning},
  author={Oreshkin, Boris and Rodríguez López, Pau and Lacoste, Alexandre},
  booktitle={Advances in Neural Information Processing Systems},
  pages={7352--7362},
  year={2018}
}

@inproceedings{raghu2020Rapid,
  title={Rapid Learning or Feature Reuse? Towards Understanding the Effectiveness of MAML},
  author={Aniruddh Raghu and Maithra Raghu and Samy Bengio and Oriol Vinyals},
  booktitle={International Conference on Learning Representations},
  year={2020},
  url={https://openreview.net/forum?id=rkgMkCEtPB}
}

@article{rajeswaran2019meta,
  title={Meta-learning with implicit gradients},
  author={Rajeswaran, Aravind and Finn, Chelsea and Kakade, Sham M and Levine, Sergey},
  journal={Advances in neural information processing systems},
  volume={32},
  year={2019}
}

@inproceedings{ravi2017optimization,
  title={Optimization as a model for few-shot learning},
  author={Ravi, Sachin and Larochelle, Hugo},
  booktitle={International conference on learning representations},
  year={2017}
}

@inproceedings{santoro2016meta,
  title={Meta-learning with memory-augmented neural networks},
  author={Santoro, Adam and Bartunov, Sergey and Botvinick, Matthew and Wierstra, Daan and Lillicrap, Timothy},
  booktitle={International conference on machine learning},
  pages={1842--1850},
  year={2016},
  organization={PMLR}
}

@inproceedings{schlag2021linear,
  title={Linear transformers are secretly fast weight programmers},
  author={Schlag, Imanol and Irie, Kazuki and Schmidhuber, J{\"u}rgen},
  booktitle={International conference on machine learning},
  pages={9355--9366},
  year={2021},
  organization={PMLR}
}

@article{schmidhuber1992learning,
  title={Learning to control fast-weight memories: An alternative to dynamic recurrent networks},
  author={Schmidhuber, J{\"u}rgen},
  journal={Neural Computation},
  volume={4},
  number={1},
  pages={131--139},
  year={1992},
  publisher={MIT Press One Rogers Street, Cambridge, MA 02142-1209, USA journals-info~…}
}

@article{shervani2023meta,
  title={Meta-learning biologically plausible plasticity rules with random feedback pathways},
  author={Shervani-Tabar, Navid and Rosenbaum, Robert},
  journal={Nature Communications},
  volume={14},
  number={1},
  pages={1805},
  year={2023},
  publisher={Nature Publishing Group UK London}
}

@article{sitzmann2020implicit,
  title={Implicit neural representations with periodic activation functions},
  author={Sitzmann, Vincent and Martel, Julien and Bergman, Alexander and Lindell, David and Wetzstein, Gordon},
  journal={Advances in neural information processing systems},
  volume={33},
  pages={7462--7473},
  year={2020}
}

@article{snell2017prototypical,
  title={Prototypical networks for few-shot learning},
  author={Snell, Jake and Swersky, Kevin and Zemel, Richard},
  journal={Advances in neural information processing systems},
  volume={30},
  year={2017}
}

@article{srivastava2014dropout,
  title={Dropout: a simple way to prevent neural networks from overfitting},
  author={Srivastava, Nitish and Hinton, Geoffrey and Krizhevsky, Alex and Sutskever, Ilya and Salakhutdinov, Ruslan},
  journal={The journal of machine learning research},
  volume={15},
  number={1},
  pages={1929--1958},
  year={2014},
  publisher={JMLR. org}
}

@inproceedings{szegedy2016rethinking,
  author    = {Christian Szegedy and Vincent Vanhoucke and
               Sergey Ioffe and Jon Shlens and Zbigniew Wojna},
  title     = {Rethinking the Inception Architecture for Computer Vision},
  booktitle = {Proceedings of the IEEE Conference on Computer Vision and Pattern Recognition},
  pages     = {2818--2826},
  year      = {2016}
}

@inproceedings{widrow1960adaptive,
  author    = {Bernard Widrow and Marcian E. Hoff},
  title     = {Adaptive Switching Circuits},
  booktitle = {1960 IRE WESCON Convention Record},
  volume    = {4},
  pages     = {96--104},
  year      = {1960}
}

@inproceedings{zintgraf2019fast,
  title={Fast context adaptation via meta-learning},
  author={Zintgraf, Luisa and Shiarli, Kyriacos and Kurin, Vitaly and Hofmann, Katja and Whiteson, Shimon},
  booktitle={International conference on machine learning},
  pages={7693--7702},
  year={2019},
  organization={PMLR}
}

\end{document}